\documentclass[letterpaper, 10 pt, conference]{ieeeconf}  
\usepackage[left=54pt, right=54.5pt, top=54.5pt, bottom=57pt]{geometry}

\IEEEoverridecommandlockouts                              

\usepackage{booktabs}
\usepackage{threeparttable}
\usepackage{cite}
\usepackage{indentfirst}
\usepackage{setspace}
\usepackage{CJK}
\usepackage{amsmath}
\usepackage{amssymb}
\usepackage{caption}
\usepackage{geometry}
\usepackage{float}
\usepackage[ruled,vlined]{algorithm2e}
\usepackage{multirow}
\usepackage{pbox}
\usepackage{here}

\usepackage{graphicx}
\usepackage{subcaption}
\usepackage{color}

\usepackage[symbol]{footmisc}

\usepackage{titlesec}

\title{\LARGE \bf
Constrained Iterative LQG for Real-Time Chance-Constrained Gaussian Belief Space Planning}

\author{Jianyu Chen$^{*}$, Yutaka Shimizu$^{*}$, Liting Sun, Masayoshi Tomizuka and Wei Zhan
\thanks{* These authors contributed equally to this work}
\thanks{
J. Chen is with the Institute for Interdisciplinary Sciences, Tsinghua University, Beijing, China, and the Shanghai Qizhi Institute, Shanghai, China. The work was conducted during J. Chen's Ph.D. study at University of California, Berkeley.}
\thanks{
Y. Shimizu is with Graduate School of Information Science and Technology, University of Tokyo, Japan. The work was conducted during Y. Shimizu's visit at University of California, Berkeley.
}
\thanks{L. Sun, M. Tomizuka and W. Zhan are with Department of Mechanical Engineering, University of California, Berkeley, USA. }

}

\begin{document}

\maketitle
\thispagestyle{empty}
\pagestyle{empty}

\begin{abstract}
Motion planning under uncertainty is of significant importance for safety-critical systems such as autonomous vehicles. Such systems have to satisfy necessary constraints (e.g., collision avoidance) with potential uncertainties coming from either disturbed system dynamics or noisy sensor measurements. 
However, existing motion planning methods cannot efficiently find the robust optimal solutions under general nonlinear and non-convex settings. In this paper, we formulate such problem as chance-constrained Gaussian belief space planning and propose the constrained iterative Linear Quadratic Gaussian (CILQG) algorithm as a real-time solution. In this algorithm, we iteratively calculate a Gaussian approximation of the belief and transform the chance-constraints. We evaluate the effectiveness of our method in simulations of autonomous driving planning tasks with static and dynamic obstacles. Results show that CILQG can handle uncertainties more appropriately and has faster computation time than baseline methods.
\end{abstract}

\section{Introduction}
When a robot working in an environment tries to accomplish a task, it will inevitably suffer from uncertainties arising in i) unmodeled or disturbed system dynamics and  ii) noisy sensor measurements. These two forms of uncertainties are common in practical robotics tasks. For example, when performing motion planning for autonomous cars, uncertainties might be introduced due to inaccurate vehicle dynamics models, localization errors, or uncertain motions of surrounding objects. Therefore, considering both dynamics and measurement uncertainties during planning is of significant importance. 

Such planning under uncertainty problem can be formally described as a partially-observable Markov decision process (POMDP)~\cite{thrun2002probabilistic}. Solving POMDPs requires planning in belief space (the set of possible states) instead of the state space, which is called belief space planning. However, general belief space planning is known to be extremely complex~\cite{papadimitriou1987complexity}. Typical solutions require discretized state and action spaces and are subject to the ``curse of dimensionality", resulting in intractable computation time. Instead of discretizing the space, Gaussian belief space planning parameterizes the beliefs as Gaussian distributions~\cite{van2017motion,van2012motion}. This body of work is promising for real-time continuous belief space planning with a running time that is polynomial in the dimension.

Moreover, in many application domains, optimizing the utility alone as in typical belief space planning methods is not enough. There are often some constraints the robot must not violate. For example, an autonomous car needs to avoid collisions with surrounding objects and constrain its control inputs within the engine limits. In belief space planning, we need to consider limiting the probability of violating constraints, which is called chance-constraint~\cite{birge2011introduction}. There are only a few works considering chance constraints in belief space planning~\cite{e2016rao,khonji2019approximability}, and they are planning in discretized space, resulting in sub-optimal plans and suffering from intractable computation time as dimension increases.

In this paper, we propose the constrained iterative LQG (CILQG) algorithm. It performs real-time Gaussian belief space planning with a general nonlinear system dynamics and measurement model while considering a general form of nonlinear and non-convex chance constraint. To solve the problem in real time, CILQG iteratively calculates Gaussian approximations of the beliefs and transforms the chance-constraints to standard linear constraints. We apply CILQG to autonomous driving trajectory planning problems with static and dynamic surrounding objects under dynamics and measurement uncertainties. The simulation results verify the performance and computation efficiency of the proposed method. 

The remainder of this paper is organized as follows. Section~\ref{sec:related} introduces related works of our work. Section~\ref{sec:problem} gives the mathematical formulation of our targeted problem. Then Section~\ref{sec:CILQG} describes the details of our proposed method. Section~\ref{sec:experiments} shows the experiments we have conducted and finally Section~\ref{sec:conclusions} concludes the paper.

\section{Related Works}\label{sec:related}
\subsection{Gaussian Belief Space Planning}
Instead of directly considering the original POMDP problem, which is in general intractable, Gaussian belief space planning finds local optimal solutions efficiently with Gaussian belief approximations and has thus become a popular trend among methods to solve POMDP. Platt et al.~\cite{platt2010belief} augmented the state with variance and used the LQG framework to find a locally-optimal control policy by assuming maximum-likelihood observations. Van den Berg et al.~\cite{van2012motion,van2017motion} approximated the belief dynamics using
an extended Kalman filter (EKF), and then used a variant of differential dynamic programming (DDP) \cite{paper:DDP-book} to plan in the belief space. Patil et al.~\cite{patil2015scaling} proposed a method to compute locally optimal plans without considering the covariance, which resulted in decreased problem dimension. Rafieisakhaei et al.~\cite{rafieisakhaei2017t} further reduced the problem dimension by restricting the policy class to linear feedback policies. Although showing impressive results for belief space planning problems, the above approaches did not formally address the constraint issue.

\subsection{Chance-Constrained Planning}
When planning under uncertainty, chance-constraint provides a formal way to account for constraints in stochastic settings. Vitus et al.~\cite{vitus2011closed} considered chance-constraints in belief space planning, but only under the linear quadratic case. The authors in \cite{da2019collision} used chance constraints to formulate the problem as a non-convex optimization problem to solve the planning problem. Okamoto et al.~\cite{okamoto2018optimalcovariance} formulated a convex optimization problem by transforming chance constraints into deterministic convex constraints. \cite{khonji2019approximability,dai2019chance} formulated a chance-constrained POMDP problem and designed a heuristic forward search algorithm to find a solution, but it only works for discrete state and action space. 

\subsection{Indirect Trajectory Optimization}
Another research area that is closely related to our work is the indirect trajectory optimization method. Although this branch of methods mainly focuses on deterministic planning problems, there are similarities between their algorithm design and ours. DDP \cite{paper:DDP-paper} \cite{paper:DDP-book} and iterative linear-quadratic regulator (ILQR) \cite{ILQR} \cite{ILQG} are the most typical algorithms for indirect trajectory optimization. They can solve the unconstrained nonlinear trajectory optimization problems efficiently by taking advantage of dynamic programming. On this basis, researchers have proposed methods to handle constraints for indirect trajectory optimization. Control-limited DDP \cite{paper:DDP-Control-Limit} considers control constraints, but it cannot solve problems with state constraints. Extended LQR \cite{paper:Iterated-LQR} \cite{paper:Extended-LQR} transforms constraints into the cost function, but cannot ensure hard constraints. \cite{paper:AL-ILQR} uses Augmented Lagrangian based optimization to solve the constrained nonlinear optimization problems. Constrained iterative LQR ~\cite{chen2017constrained}~\cite{chen2019autonomous} handles state and control input constraints using barrier function to transform constraints in a way similar to interior-point method. Therefore, CILQR can be applied to general nonlinear systems with nonlinear constraints. 

\section{Problem Formulation}\label{sec:problem}
We now define the problem we will discuss in this paper. Let $\cal X\subset$ $\mathbb{R}^n$ be the space of all possible states $x$ of the agent, $\cal U\subset$ ${\mathbb R}^m$ be the space of of all executable control commands $u$ of the agent, and $\cal Y\subset$ ${\mathbb R}^r$ be the space of all possible sensor measurements $y$ of the agent. We consider a generic form of nonlinear stochastic system dynamics and measurement model:
\begin{equation}
\begin{aligned}
    x_{k+1}&=f\left(x_k,u_k,w_k\right),\qquad w_k\sim{\cal N}\left(0,\Sigma_w\right)\\
    y_{k+1}&=h\left(x_{k+1},v_{k+1}\right),\qquad v_{k+1}\sim{\cal N}\left(0,\Sigma_v\right)
\end{aligned}
\end{equation}
where $x_k$, $u_k$, and $y_k$ are the state, control input, and sensor measurement of the agent at time step $k$. $w_k$ represents the system noise and $v_k$ represent the measurement noise, both are assumed Gaussian distributions with zero mean. $\Sigma_w$ and $\Sigma_v$ represent their variance. Note that although the noise terms here are Gaussian, the distribution of $x_{k+1}$ and $y_{k+1}$ can be non-Gaussian since $w_k$ and $v_k$ will go through the nonlinear transformations $f$ and $h$.

To plan under uncertainty, we formulate the problem as a POMDP or belief space planning problem, where the belief of the agent is defined as the distribution of the state conditioned on historical measurements and control inputs: 
\begin{equation}
    b_k=\texttt{Pr}\left(\left.x_k\right|y_{1:k},u_{0:k-1}\right)
\end{equation}

Here we assume the belief is represented by a Gaussian distribution in this paper, which is described by the mean and variance of the state $b_k=\left(\mu_k,\Sigma_k\right)$. Our goal is to find a control policy $u_k=\pi_k\left(b_k\right)$ that minimizes the cost function:
\begin{equation}
    \min_{u_{0:N-1}}\;{\mathbb E}\left[l^N(b_N)+\sum_{k=0}^{N-1}l^k(b_k,u_k)\right]
\end{equation}
where $l^k(b_k,u_k)$ is a general nonlinear stage cost function at time step $k$ and $l^N(b_N)$ is a general nonlinear terminal cost function. 

In addition to minimizing the cost, we also consider the chance-constraints on the states and control inputs:
\begin{equation}
\begin{aligned}
    \texttt{Pr}(g_x^k(x_k) \leq 0) \geq p,\qquad\texttt{Pr}(g_u^k(u_k) \leq 0) \geq p
\end{aligned}
\end{equation}
where $g_x^k(x_k) \leq 0$ and $g_u^k(u_k) \leq 0$ represents the nonlinear state and control input constraints at time step $k$, and $p$ is the chance-constraint threshold, which should be greater than $0.5$. As a summary, the targeting problem for our paper is formulated as the following:
\begin{subequations}
\label{equation:cilqg-formulation}
\begin{gather}
    \label{equation:cilqg-formulation-cost}
    \min_{u_{0:N-1}}\;{\mathbb E}\left[l^N(b_N)+\sum_{k=0}^{N-1}l^k(b_k,u_k)\right] \\
    \label{equation:cilqg-formulation-dynamics}
    x_{k+1}=f\left(x_k,u_k,w_k\right),\qquad w_k\sim{\cal N}\left(0,\Sigma_w\right)\\
    \label{equation:cilqg-formulation-measurement}
    y_{k+1}=h\left(x_{k+1},v_{k+1}\right),\qquad v_{k+1}\sim{\cal N}\left(0,\Sigma_v\right)\\
    \label{equation:cilqg-formulation-state-chance}
    \texttt{Pr}(g_x^k(x_k) \leq 0) \geq p \\
    \label{equation:cilqg-formulation-control-chance}
    \texttt{Pr}(g_u^k(u_k)\leq 0) \geq p\\
    \label{equation:cilqg-formulation-initial-belief}
    b_0=\left(\mu_0,\Sigma_0\right)
\end{gather}
\end{subequations}
where $\mu_0=x_0$ is initial mean state, $\Sigma_0$ is initial state covariance and~\eqref{equation:cilqg-formulation-initial-belief} represents initial belief of the problem.

\section{Constrained Iterative LQG}\label{sec:CILQG}
To solve the chance-constrained Gaussian belief space planning problem formulated in Section~\ref{sec:problem} in real time, we propose the constrained iterative LQG (CILQG) algorithm. CILQG first linearizes the nonlinear stochastic system. Then based on the linearized system, it propagates the belief using a modified Kalman filter. After that, it transforms the chance-constraint into a linear inequality constraint. The algorithm then iterates in an outer-inner loop form similar to~\cite{chen2019autonomous}. We describe details of the algorithm in this section.

\subsection{System Linearization and Belief Dynamics}\label{sec:belief_dynamics}
To make the algorithm tractable, we first need to linearize the nonlinear stochastic system dynamics as well as the nonlinear measurement model~\eqref{equation:cilqg-formulation-dynamics} \eqref{equation:cilqg-formulation-measurement}. Both dynamics and measurement model are linearized around a nominal trajectory $\bar{x}, \bar{y}, \bar{w}=0, \bar{v}=0$:
\begin{equation}\label{eq:linearize}
    \begin{aligned}
        x_{k+1}
        &\approx \bar{x}_{k+1}+A_k\left(x_k-\bar{x}_k\right)+B_k\left(u_k-\bar{u}_k\right)+W_k w_k\\
        y_{k+1}
        &\approx h\left(\bar{x}_{k+1},0\right)+H_{k+1}\left(x_{k+1}-\bar{x}_{k+1}\right)+V_{k+1} v_{k+1}
    \end{aligned}
\end{equation}
where
\begin{equation}\label{eq:coefficient}
\begin{aligned}
    A_k&=\frac{\partial f\left(\bar{x}_k,\bar{u}_k,0\right)}{\partial x},\quad B_k=\frac{\partial f\left(\bar{x}_k,\bar{u}_k,0\right)}{\partial u},\\
    W_k&=\frac{\partial f\left(\bar{x}_k,\bar{u}_k,0\right)}{\partial w},\\ H_{k+1}&=\frac{\partial h\left(\bar{x}_{k+1},0\right)}{\partial x},\quad V_{k+1}=\frac{\partial h\left(\bar{x}_{k+1},0\right)}{\partial v}
\end{aligned}
\end{equation}

The model is linearized around the nominal trajectory $\left\{\bar{x}_k,\bar{u}_k\right\}$ and zero means of the noises $w_k$ and $v_k$. With this linearized model, we can apply Kalman filter to obtain the belief dynamics. Specifically, we can write the prior update as:
\begin{equation}\label{eq:KF_prior}
\begin{aligned}
    \hat{x}_{k+1}^p &= \bar{x}_{k+1}+A_k\left(\hat{x}_k^m-\bar{x}_k\right) + B_k \left(u_{k}-\bar{u}_k\right) \\
    \hat{\Sigma}_{k+1}^p &= A_k\hat{\Sigma}_k^m A_k^\top + W_k\Sigma_w W_k^\top
\end{aligned}
\end{equation}
where $\left(\hat{x}_k^m, \hat{\Sigma}_k^m\right)$ represents the posterior estimation, and $\left(\hat{x}_{k+1}^p, \hat{\Sigma}_{k+1}^p\right)$ represents the prior estimation of the Gaussian belief. The measurement update can be written as:
\begin{equation}\label{eq:KF_measurement}
\begin{aligned}
    K_{k+1}&=\hat{\Sigma}_{k+1}^p H_{k+1}^\top\left(H_{k+1}\hat{\Sigma}_{k+1}^p H_{k+1}^\top+V_{k+1}\hat{\Sigma}_k^m V_{k+1}^\top\right)^{-1}\\
    \hat{x}_{k+1}^m&=\hat{x}_{k+1}^p+K_{k+1}\left(y_{k+1}-\right.\\
    &\qquad\qquad\left.\left(h\left(\bar{x}_{k+1},0\right)+H_{k+1}\left(\hat{x}_{k+1}^p-\bar{x}_{k+1}\right)\right)\right)\\
    \hat{\Sigma}_{k+1}^m &= \left(I-K_{k+1}H_{k+1}\right)\hat{\Sigma}_{k+1}^p
\end{aligned}
\end{equation}
where $K_{k+1}$ is the Kalman filter gain. Since we do not know the value of future measurement $y_{k+1}$ at time step $k$, we treat it as the output of the measurement function and approximate it with its expectation:
\begin{equation}\label{eq:fake_measurement}
\begin{aligned}
    y_{k+1}&\approx{\mathbb E}\left[h\left(\bar{x}_{k+1},0\right)+H_{k+1}\left(x_{k+1}-\bar{x}_{k+1}\right)+V_{k+1} v_{k+1}\right]\\
    &=h\left(\bar{x}_{k+1},0\right)+H_{k+1}\left(\hat{x}_{k+1}^p-\bar{x}_{k+1}\right)
\end{aligned}
\end{equation}

Substitute \eqref{eq:fake_measurement} into \eqref{eq:KF_measurement} we have $\hat{x}_{k+1}^m=\hat{x}_{k+1}^p$. Therefore, given the initial belief $\left(\hat{x}_0^m,\hat{\Sigma}_0^m\right)=\left(\mu_0,\Sigma_0\right)$ we can propagate the mean of the belief with the system dynamics:
\begin{equation}
    \hat{x}_{k+1}^m=\bar{x}_{k+1}+A_k\left(\hat{x}_k^m-\bar{x}_k\right) + B_k \left(u_{k}-\bar{u}_k\right)
\end{equation}
and propagate the variance of the belief following \eqref{eq:KF_prior} and \eqref{eq:KF_measurement}. Note that the variance update process is dependent only on the system coefficients but not on the states or actions. The variance propagation algorithm is summarized in Algorithm~\ref{alg:belief}.
\begin{algorithm}
    \SetAlgoLined
    \SetNoFillComment
    \SetKwData{Left}{left}\SetKwData{This}{this}\SetKwData{Up}{up}
    \SetKwFunction{Union}{Union}\SetKwFunction{FindCompress}{FindCompress}
    \SetKwInOut{Input}{input}\SetKwInOut{Output}{output}
    \Input{Nominal trajectory $\left(\bar{\mathbf{x}},\bar{\mathbf{u}}\right)$; Initial belief $\left(\hat{x}_0^m,\hat{\Sigma}_0^m\right)$}
    \Output{Variance sequence $\hat{\mathbf{\Sigma}}^m$}
    \For{$k=0$ \KwTo $N-1$}
    {
        Linearize the system to obtain $A_k$, $B_k$, $W_k$, $H_{k+1}$ and $V_{k+1}$ following \eqref{eq:linearize} and \eqref{eq:coefficient};\\
        Obtain the prior estimation of variance $\hat{\Sigma}_{k+1}^p$ using \eqref{eq:KF_prior};\\
        Obtain the posterior estimation of variance $\hat{\Sigma}_{k+1}^m$ using \eqref{eq:KF_measurement};\\
        $\hat{\mathbf{\Sigma}}^m \leftarrow$ append$\left(\hat{\Sigma}_{k+1}^m\right)$
    }
    \Return $\hat{\mathbf{\Sigma}}^m$
    \caption{Variance Propagation}
    \label{alg:belief}
\end{algorithm}

\subsection{Handling Chance-Constraints}
In general, directly considering the original formulation of chance-constraints~\eqref{equation:cilqg-formulation-state-chance}~\eqref{equation:cilqg-formulation-control-chance} are extremely challenging. In this section, we describe how to transform complex chance-constraints into simple deterministic constraints. We will only discuss the state chance-constraint~\eqref{equation:cilqg-formulation-state-chance} here without loss of generality, as the control input chance-constraint~\eqref{equation:cilqg-formulation-control-chance} follows the same procedure.

By linearizing the constraint function $g_x^k\left(x_k\right)\leq 0$ around the nominal state $\bar{x}_k$ we get:
\begin{equation}\label{eq:linearize_constraint1}
    G_x^k x_k + m_x^k \leq 0
\end{equation}
where
\begin{equation}\label{eq:linearize_constraint2}
    G_x^k = \frac{\partial g_x^k\left(\bar{x}_k\right)}{\partial x},\quad m_x^k=\bar{x}_k-G_x^k\bar{x}_k
\end{equation}

We further decompose $x_k$ into a deterministic component $z_k \in R^n$ and a stochastic component $e_k \in R^n$ as the following:
\begin{equation} \label{equation:state-decomposition}
    x_k = z_k + e_k
\end{equation}
where $e_k\sim {\cal N}\left(0,\hat{\Sigma}_k^m\right)$ is a zero mean Gaussian. We then get a simplified chance-constraint:
\begin{equation} \label{equation:chance-constraint-decomposition}
    \texttt{Pr}\left(G_x^k z_k + G_x^k e_k + m_x^k \leq 0\right) \geq p
\end{equation}
This is equal to the following constraints \cite{Chace_transform} \cite{SMPC_Chance}:
\begin{subequations}
\begin{gather}
    \label{eq:linear_cons}
    G_x^k z_k \leq -m_x^k - \gamma_x^k \\
    \texttt{Pr}\left(G_x^k e_k \leq \gamma_x^k\right) = p
\end{gather}
\end{subequations}

Since $G_x^k e_k\sim{\cal N}\left(0,G_x^k\hat{\Sigma}_k^m \left(G_x^k\right)^T\right)$, we can calculate the value of $\gamma_x^k$ analytically:
\begin{equation} \label{equation:calculate-gamma}
    \gamma_x^k = \sqrt{2G_x^k\hat{\Sigma}_k^m \left(G_x^k\right)^T} erf^{-1}(2p-1)
\end{equation}
where $erf^{-1}\left(\cdot\right)$ is the inverse error function. Therefore, the original chance-constraint is transformed to a linear constraint~\eqref{eq:linear_cons} with $\gamma_x^k$ calculated by~\eqref{equation:calculate-gamma}. The chance-constraint transformation algorithm is summarized in Algorithm~\ref{alg:chance}.
\begin{algorithm}
    \SetAlgoLined
    \SetNoFillComment
    \SetKwData{Left}{left}\SetKwData{This}{this}\SetKwData{Up}{up}
    \SetKwFunction{Union}{Union}\SetKwFunction{FindCompress}{FindCompress}
    \SetKwInOut{Input}{input}\SetKwInOut{Output}{output}
    \Input{Nominal state $\bar{x}_k$;\\
    State variance $\hat{\Sigma}_k^m$}
    \Output{Transformed linear constraint function coefficients $G_x^k$, $m_x^k$, $\gamma_x^k$}
    Linearize the constraint function following \eqref{eq:linearize_constraint1} \eqref{eq:linearize_constraint2} and obtain the coefficients $G_x^k$, $m_x^k$;\\
    
    Calculate the coefficient $\gamma_x^k$ according to \eqref{equation:calculate-gamma};\\
    
    \Return $G_x^k$, $m_x^k$, $\gamma_x^k$
    \caption{Chance-Constraint}
    \label{alg:chance}
\end{algorithm}

Let's now take a look at how we define the constraint functions $g_x^k$ and $g_u^k$. For control input constraint, we define a box constraint $\underline{u}\leq u_k\leq \overline{u}$ where $\underline{u}$ is the lower bound and $\overline{u}$ is the upper bound of the control input. For state constraint, we focus on collision avoidance constraints, which are defined in the following two ways:

\subsubsection{Constraints considering obstacles' shapes}\label{sec:shape}
\begin{figure}
    \centering
    \includegraphics[width=0.45\textwidth]{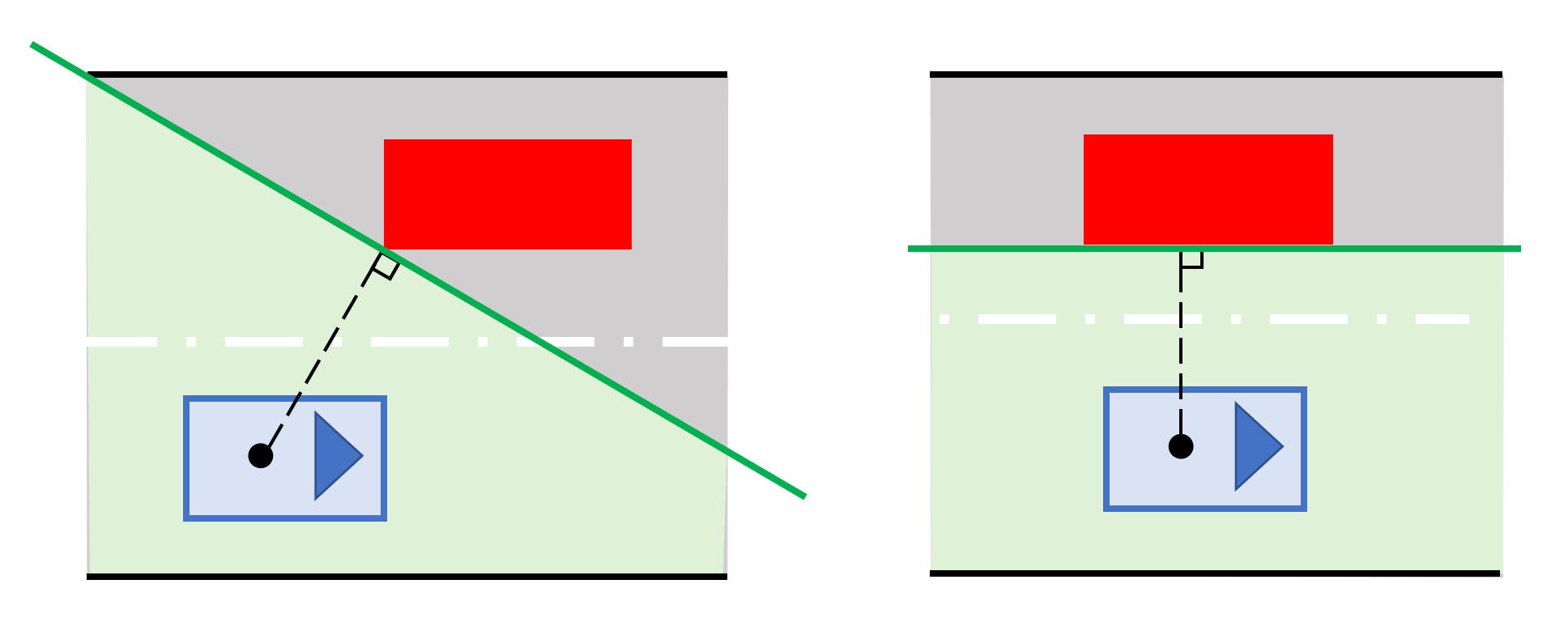}
    \caption{Illustration of convex feasible set calculation}
    \label{fig:convex_feasible_set}
\end{figure}

 Now define $\Gamma_i=\left\{x:\phi_i\left(x\right)\geq 0\right\}$ to be the space outside of the $i$th obstacle, where $\phi_i$ is the signed distance function to the boundary of $i$th obstacle. Since in general $\Gamma$ can be highly non-convex, to make the downstream optimization problem tractable we instead calculate a convex feasible set~\cite{changliu2017theconvexfeasible,chen2018foad} of $\Gamma_i$:
\begin{equation}
    {\cal F}_i\left(\bar{x}_k\right)=\left\{x:\phi_i\left(\bar{x}_k\right)+\frac{\partial\phi_i\left(\bar{x}_k\right)}{\partial x}\left(x-\bar{x}_k\right)\geq 0\right\}
\end{equation}
where $\bar{x}_k$ is the agent's nominal state. When the obstacle's shape is a convex polygon (e.g, a rectangle), we can have a rather intuitive explanation of the convex feasible set. Fig.\ref{fig:convex_feasible_set} shows an example of convex feasible set calculation for on-road autonomous driving. The blue rectangle represents the ego vehicle and the red rectangle represents a surrounding obstacle. The green region is the corresponding convex feasible set, which is a half space with its boundary perpendicular to the closest connecting line from the center of ego vehicle to the obstacle polygon.

In this paper we will only consider the cases where the obstacles have convex polygon shapes. Therefore ${\cal F}_i\left(\bar{x}_k\right)$ is always a half space. When there are multiple obstacles, we calculate the intersection of all convex feasible sets ${\cal F}=\bigcap_i{\cal F}_i$, which is a combination of a group of linear (half space) constraints and is still convex. It is shown in \cite{changliu2017theconvexfeasible} that ${\cal F}$ is non-empty if the obstacles are disjoint.

\subsubsection{Constraints considering obstacles' uncertainties}\label{sec:uncertainty}
\begin{figure}
    \centering
    \includegraphics[width=0.48\textwidth]{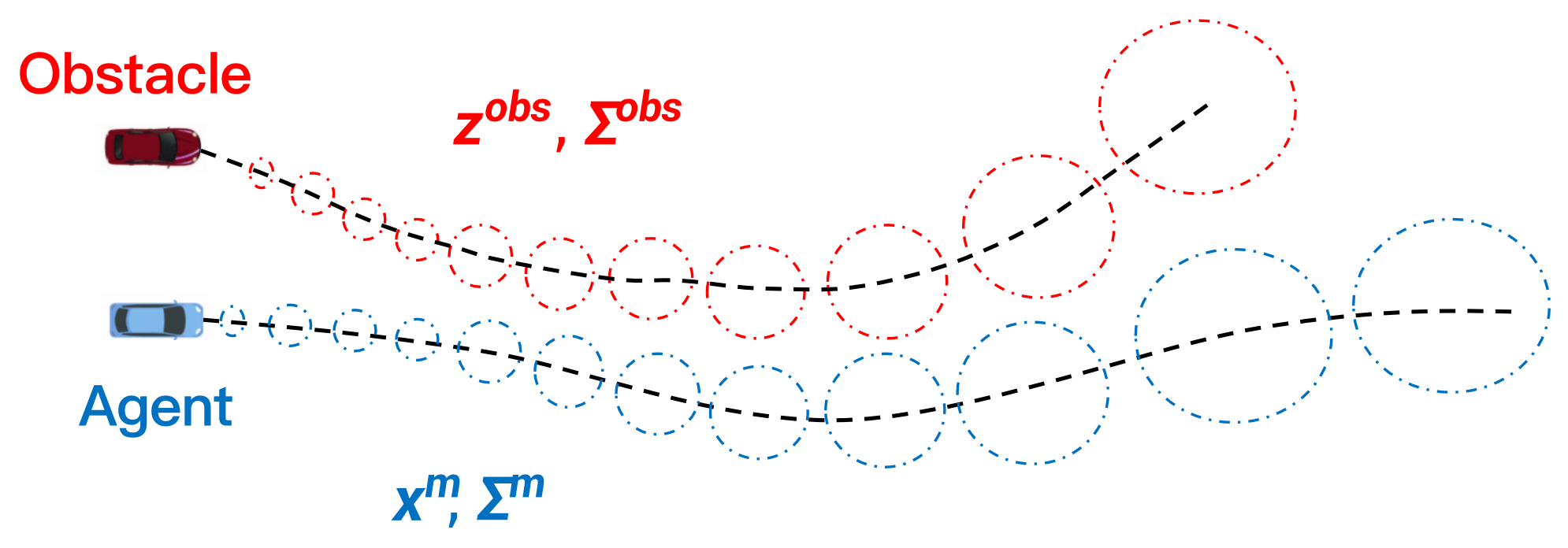}
    \caption{Considering obstacle's uncertainty}
    \label{fig:dynamic_obstacle_avoidance}
\end{figure}

There are always uncertainties arising from detection, localization, and prediction for surrounding obstacles, especially for moving ones. It's better if we can consider uncertainties with not only the ego agent's motion but also the obstacles', as shown in Fig.\ref{fig:dynamic_obstacle_avoidance}. Let $x_k^{\textrm{obs}}$ represents the estimated state of the obstacle, which is stochastic and can be decoupled the same way as done in~\eqref{equation:state-decomposition}:
\begin{equation}
    x_k^{\textrm{obs}} = z_k^{\textrm{obs}} + e_k^{\textrm{obs}}
\end{equation}
where $e_k^{\textrm{obs}}$ is a zero mean Gaussian random variance with variance $\Sigma_k^\textrm{obs}$, and $z_k^{\textrm{obs}}$ is the mean state, which is considered fixed during the planning process. The constraint function is then written as:
\begin{equation}
    \eta-\left\|x_k-x_k^{\textrm{obs}}\right\| \leq 0
\end{equation}
where $\eta$ is a safety margin. Note that now the variance of $e_k$ in \eqref{equation:state-decomposition} should be $\hat{\Sigma}_k^m+\Sigma_k^\textrm{obs}$, since we need to consider the obstacle's uncertainty together with the agent's uncertainty.

\subsection{Constrained Iterative LQG Algorithm}\label{sec:ilqr}
We have introduced how to calculate the belief dynamics and how to handle the chance constraints. Let's take a look at our proposed Constrained Iterative LQG algorithm, which is summarized in Algorithm~\ref{alg:cilqg}. The algorithm is designed in an outer-inner loop framework. The outer-loop calculates the estimated state variance and transforms the chance constraint, which is then augmented to the cost function. Then the inner-loop performs iterative LQR to update the control input sequence as well as the trajectory. Details of the cost augmentation procedure and the applied iterative LQR algorithm can be found in~\cite{chen2019autonomous}, here we briefly introduce the ILQR backward pass used in our algorithm. 

According to equation~\eqref{eq:linearize}, we can obtain $A_k$ and $B_k$ by linearizing the system dynamics around the nominal trajectory $\left(\bar{\textbf{x}}, \bar{\textbf{u}}\right)$.
Now quadratize the cost function:
\begin{equation}
\begin{aligned}
    l\left(x_k,u_k\right)\approx \Tilde{x}_k^\top l_x^k+\Tilde{u}_k^\top l_u^k+\frac{1}{2}\Tilde{x}_k^\top l_{xx}^k\Tilde{x}_k+\frac{1}{2}\Tilde{u}_k^\top l_{uu}^k\Tilde{u}_k\\
    +\Tilde{u}_k^\top l_{ux}^k\Tilde{x}_k+l\left(\bar{x}_k,\bar{u}_k\right)
\end{aligned}
\end{equation}
where $\Tilde{x}_k=x_k-\bar{x}_k$ and $\Tilde{u}_k=u_k-\bar{u}_k$ and subscripts denote the Jacobians and Hessians of the cost function. Note that our cost design omits the variance part during our implementation because our estimated variance of the state does not depend on the control inputs as shown in Section~\ref{sec:belief_dynamics}. Thus the variance does not enter the optimization procedure. Then recursively estimates the Q-function from backward:
\begin{equation} \label{equation:backward-definition}
    \begin{aligned}
    Q_{xx}^k&=l_{xx}^k+A_k^\top V_{xx}^{k+1}A_k \qquad Q_x^k=l_x^k+A_k^\top V_x^{k+1}\\
    Q_{ux}^k&=l_{uu}^k+B_k^\top V_{xx}^{k+1}A_k \qquad Q_u^k=l_u^k+B_k^\top V_x^{k+1}\\
    Q_{uu}^k&=l_{uu}^k+B_k^\top V_{xx}^{k+1}B_k + \rho I
    \end{aligned}
\end{equation}
as well as the value function and linear policy terms:
\begin{equation}
    \begin{aligned}
    V_x^k&=Q_x^k-Q_{ux}^\top \left(Q_{uu}^k\right)^{-1}Q_u^k\\
    V_{xx}^k&=Q_{xx}^k-Q_{ux}^\top \left(Q_{uu}^k\right)^{-1}Q_{ux}^k \left(Q_{uu}^k\right)^{-1}Q_u^k
    \end{aligned}
\end{equation}

Then the optimal control input is given by:
\begin{equation}
    u_k^*=\bar{u}_k-\left(Q_{uu}^k\right)^{-1}\left(Q_u^k+Q_{ux}^k\left(x_k-\bar{x}_k\right)\right)
\end{equation}
Note that, we add regularization term $\rho$ in the equation~\eqref{equation:backward-definition} to guarantee $Q_{uu}^k$ is invertible.

\begin{algorithm}
    \SetAlgoLined
    \SetNoFillComment
    \SetKwData{Left}{left}\SetKwData{This}{this}\SetKwData{Up}{up}
    \SetKwFunction{Union}{Union}\SetKwFunction{FindCompress}{FindCompress}
    \SetKwInOut{Input}{input}\SetKwInOut{Output}{output}
    \Input{Feasible initial control sequence $\bar{\mathbf{u}}$; \hspace{2mm} $t:=t^{(0)}>0, \hspace{2mm} \mu > 1, \hspace{2mm}1\geq\alpha\geq 0$; \\ Initial belief $\left(\hat{x}_0^m,\hat{\Sigma}_0^m\right)$;\\ Chance-constraint threshold $p$}
    \Output{Optimal control sequence $\bar{\mathbf{u}}^*$ and corresponding belief trajectory $\left(\hat{\mathbf{x}}^m,\hat{\mathbf{\Sigma}}^m\right)$}
    $\bar{\mathbf{x}}\leftarrow$ Forward simulation with system dynamics~\eqref{equation:cilqg-formulation-dynamics} under zero noises using $\bar{\mathbf{u}}$;\\
    \While(\tcc*[h]{Outer Loop}){not converge}
    {
        $\hat{\mathbf{\Sigma}}^m\leftarrow$
        \texttt{VariancePropagation}$\left(\bar{\mathbf{x}},\bar{\mathbf{u}},\hat{x}_0^m,\hat{\Sigma}_0^m\right)$ \\
        
        \For{$k=0$ \KwTo $N-1$}
        {
        $G_x^k, m_x^k, \gamma_x^k\leftarrow$
        \texttt{StateChanceConstraint}$\left(\bar{x}_k, \hat{\Sigma}_k^m\right)$;\\
        $G_u^k, m_u^k, \gamma_u^k\leftarrow$
        \texttt{ControlChanceConstraint}$\left(\bar{u}_k, \Sigma_u\right)$;\\
        $l^k\left(b_k,u_k\right)\leftarrow l^k\left(b_k,u_k\right)-\frac{1}{t} \log{(-G_x^k x_k -m_x^k-\gamma_x^k)}-\frac{1}{t} \log{(-G_u^k u_k-m_u^k-\gamma_u^k)}$;\\
        }
        \While(\tcc*[h]{Inner Loop}){not converge}
        {
            Compute control sequence $\mathbf{u}^*$ by performing an ILQR backward pass described in Section~\ref{sec:ilqr};\\
            \While(\tcc*[h]{Line Search}){cost increased or constraints violated}
            {
                $\bar{\mathbf{u}}:=\alpha\mathbf{u}^*$;\\
                $\bar{\mathbf{x}}\leftarrow$ Forward simulation with system dynamics~\eqref{equation:cilqg-formulation-dynamics} under zero noises using $\bar{\mathbf{u}}$;\\
            }
        }
        $\bar{\mathbf{u}}^*:=\bar{\mathbf{u}}$;\\
        $\hat{\mathbf{x}}^m:=\bar{\mathbf{x}}$; \\
        $t:=\mu t$; \\
    }
    \caption{CILQG Algorithm}
    \label{alg:cilqg}
    \Return $\bar{\mathbf{u}}^*, \hat{\mathbf{x}}^m, \hat{\mathbf{\Sigma}}^m$
\end{algorithm}

\section{Experiments}\label{sec:experiments}
We evaluate the performance of the proposed method on simulations of autonomous driving motion planning tasks in the presence of both static and dynamic obstacles. Furthermore, we compare our method with the following three related approaches:
\begin{itemize}
    \item{\textbf{CILQR}:} Constrained iterative LQR~\cite{chen2019autonomous} is a deterministic motion planning algorithm for autonomous driving, which does not consider the measurement model and the system noise.
    \item{\textbf{Gaussian Belief Space Planning(GBSP)}:} This is a soft-constraint version of our method, which shares the same problem formulation and similar methodology with existing Gaussian belief space planning approaches~\cite{van2012motion,van2017motion}.
    \item{\textbf{Open CILQG}:} This is an open-loop version of our method, which does not consider the measurement model and thus does not use filtering techniques to make a closed-loop estimation of the belief. The problem formulation is similar to~\cite{okamoto2018optimalcovariance} but for nonlinear dynamics.
\end{itemize}

\begin{figure}
  \centering
  \includegraphics[scale=0.32]{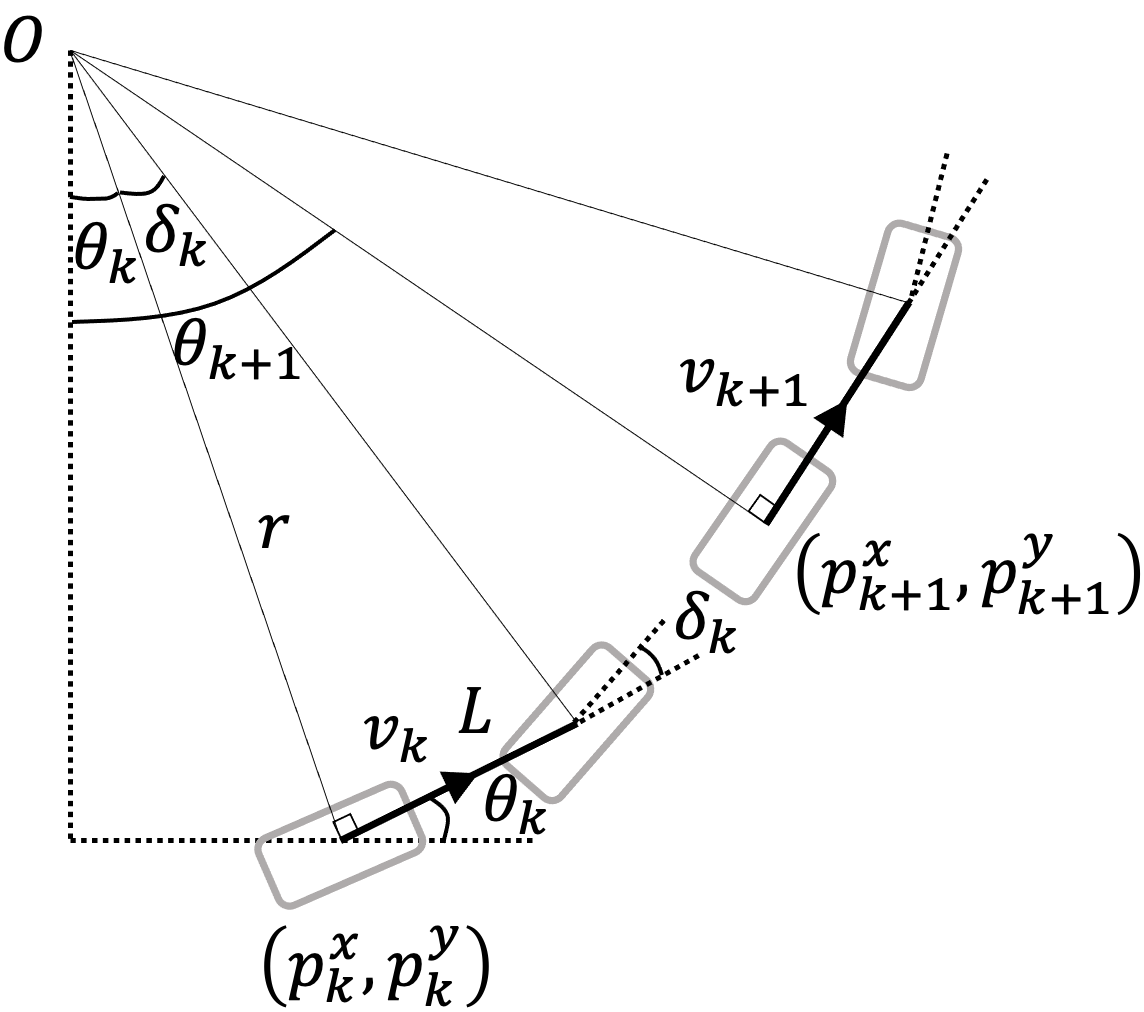}
  \caption{\label{fig:model}The vehicle bicycle kinematic model}
\end{figure}

The vehicle model we use throughout the experiments is the bicycle kinematics model, which is shown in Fig.~\ref{fig:model}.
The vehicle state vector $x_k$ at the current time step $k$ includes the 2D position $\left( {{p^x_k},\;{p^y_k}} \right)$, the velocity ${v_k}$ and the heading ${\theta _k}$. The control input vector $u_k$ includes the acceleration $a_k$ and the steering angle $\delta_k$. $L$ is the wheel base. Since for digital control systems the control input will maintain the same in a sampling time $T_r$, the vehicle will rotate around the instant center $O$ with rotation radius $r$. The distance the vehicle moves in one sampling time is $d = {v_k}{T_r} + \frac{1}{2}aT_r^2$ and the curvature is $\kappa  = \frac{{\tan \delta }}{L}$. We assume there are noises $w_a$ and $w_k$ inserted to the acceleration and curvature. Therefore, the vehicle system dynamics can be written as:
\begin{align}
\label{eq:model}
\begin{array}{l}
{v_{k+1}} = {v_k} + \left(a+w_a\right){T_r}\\
{\theta_{k+1}} = {\theta _k} + \int_0^d {\left(\kappa+w_k\right) ds}\\
{p^x_{k+1}} = {p^x_k} + \int_0^d {\cos \left( {{\theta _k} + \left(\kappa+w_k\right) s} \right)ds}\\
{p^y_{k+1}} = {p^y_k} + \int_0^d {\sin \left( {{\theta _k} + \left(\kappa+w_k\right) s} \right)ds}
\end{array}
\end{align}

The measurement model is set as the true state plus a noise proportional to the velocity (The velocity is assumed always positive):
\begin{equation}
    \begin{aligned}
    y_k=x_k+v_k m_k,\qquad v_{k+1}\sim{\cal N}\left(0,\Sigma_v\right)
    \end{aligned}
\end{equation}
This form of measurement model is based on the assumption that the sensor measurement becomes inaccurate as the speed increases. In this simulation, we use the following cost function~\eqref{equation:cilqg-formulation-cost}:
\begin{equation}\label{equation:test_cost_function}
\begin{aligned}
    l^k\left(b_k, u_k \right)&=(b_k- x_{k, ref})^\top Q(b_k - x_{k, ref}) + u_k^\top Ru_k \\
    l^N\left(b_N \right)&=(b_N- x_{N, ref})^\top Q_f(b_N - x_{N, ref})
\end{aligned}
\end{equation}
where $x_{k,ref}$ is a reference trajectory generated by a higher-level global trajectory planner. The Matrix $Q, Q_f$ and $R$ are coefficients determining the shape of the cost. Note that although we use this specific quadratic cost function in our experiments, it is also possible to give general nonlinear costs in other settings. The ILQR step illustrated in Section~\ref{sec:ilqr} will handle the nonlinearity by quadratize the cost function.

We set the chance-constraint threshold $p$ as 0.98 and the prediction horizon as $N=50$ for all experiments. In addition, we use $T_r = 0.2$ as sampling time in this simulation. The simulation is implemented in C++ on a desktop PC with 3.10GHz Intel Core i9-9900 CPU. Details of the experiment results are introduced in the following subsections.

\subsection{Static Obstacle Avoidance}
We first test our method in two scenarios with static obstacles. In this case, we use the constraints considering obstacles' shapes as described in Section~\ref{sec:shape}. 

\begin{figure}
    \centering
    \includegraphics[scale=0.28]{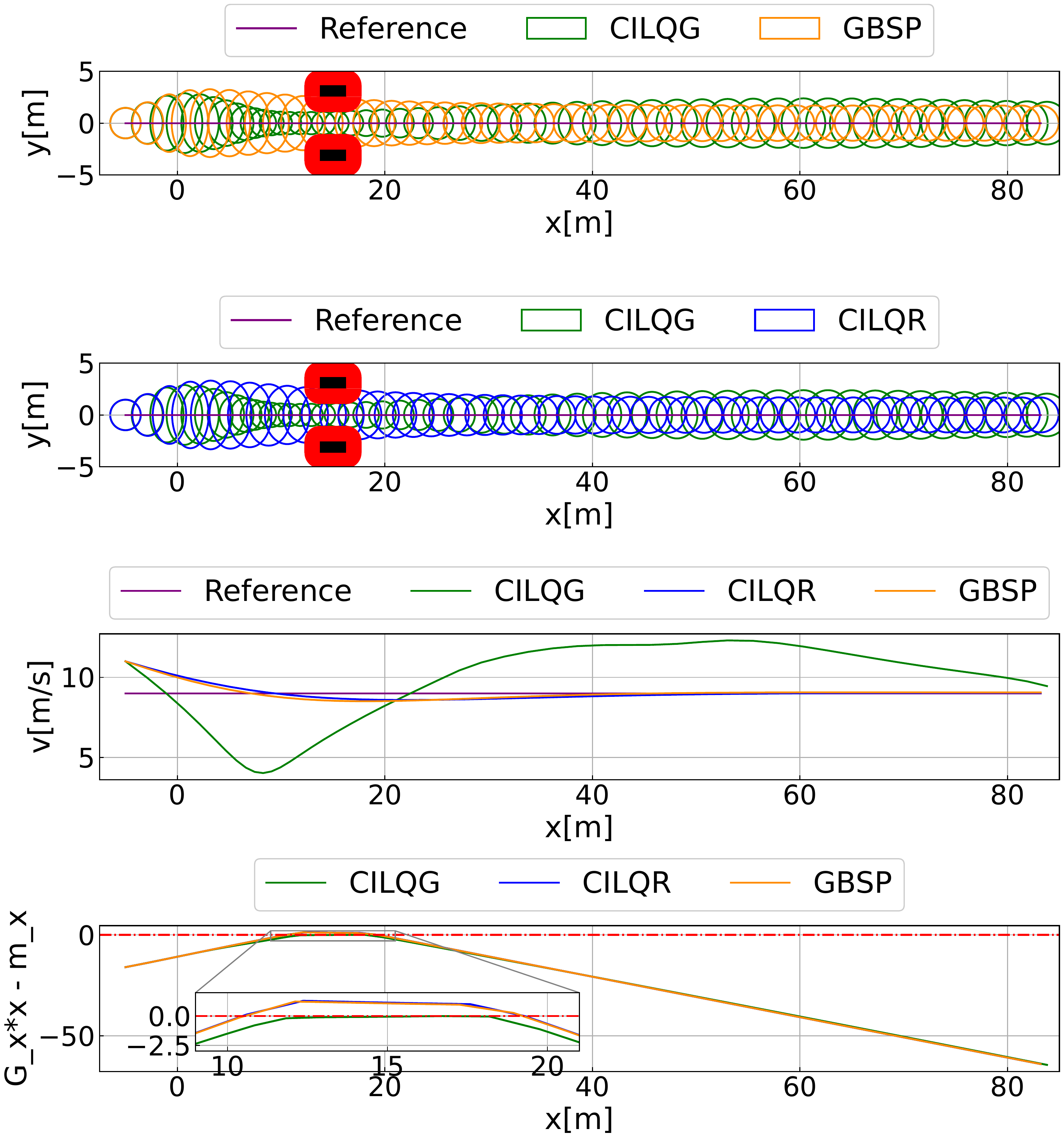}
    \caption{CILQG with two static obstacles}
    \label{fig:experiment_with_two_static_obstacles}
\end{figure}

First, our vehicle is required to pass through the interval between two static obstacles safely and efficiently. Fig.\ref{fig:experiment_with_two_static_obstacles} shows the simulation results. The top two sub-figures show the planning results of CILQG, GBSP, and CILQR, represented by different colors. The ellipse represents the confidence region with probability $p=0.98$, which is equal to the chance-constraint threshold. Intuitively, the confidence region indicates where the vehicle is probable to be positioned. The dark red rectangles represent the static obstacles, and the light red regions indicate the safety margins. We can see that the result of CILQG keeps its confidence region collision-free with the obstacles, while other methods fail. The third sub-figure shows the speed profile of the three methods. We can find that CILQG decelerates before passing through the obstacles to decrease the uncertainty and then accelerates to keep its driving efficiency. The bottom sub-figure shows the left-hand side of the constraint function~\eqref{eq:linearize_constraint1}, which should ideally be non-positive to keep the state chance-constraints satisfied. We can see that only our proposed CILQG succeeds.

\begin{figure}
    \centering
    \includegraphics[width=0.45\textwidth]{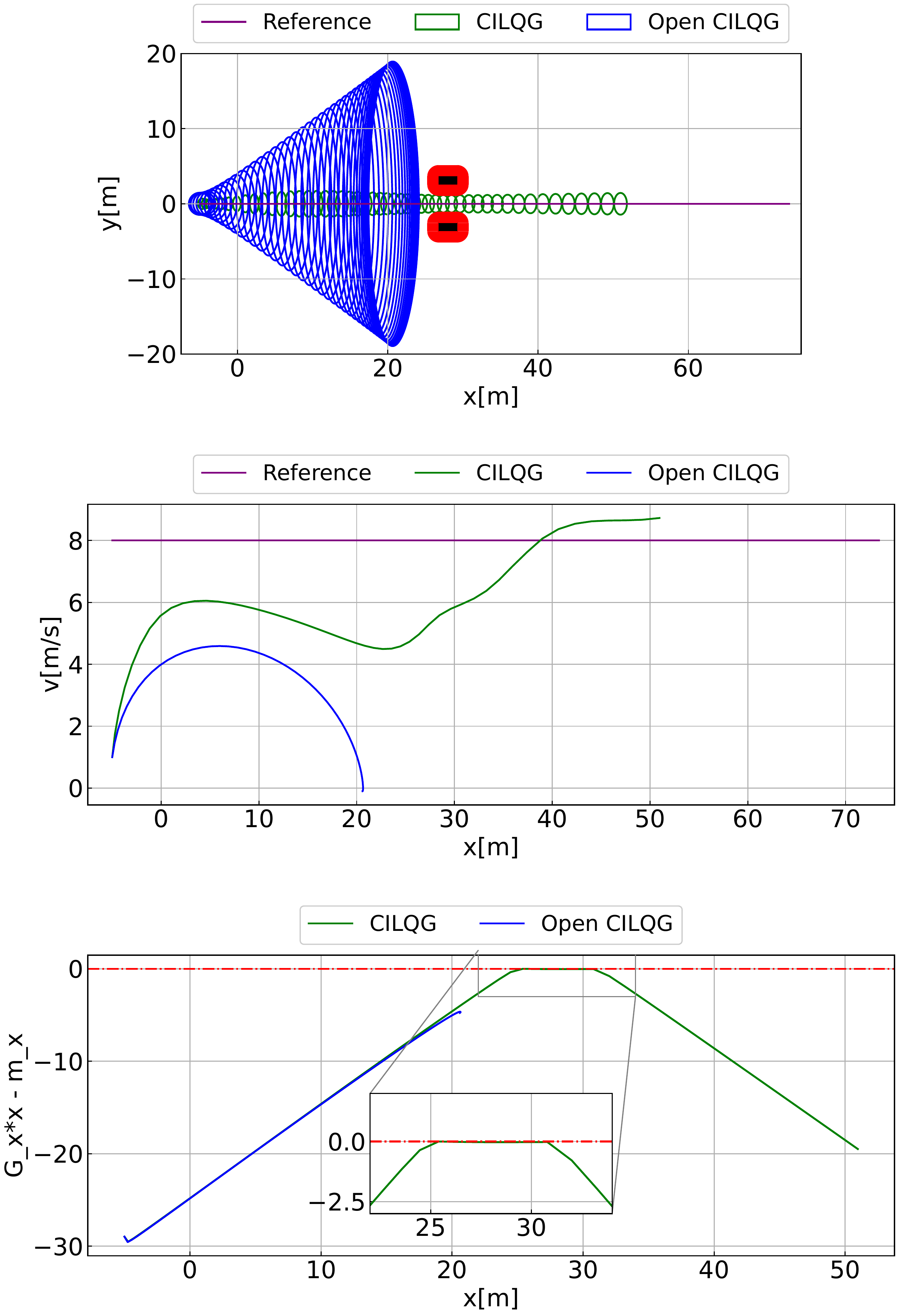}
    \caption{Closed-loop and Open-loop CILQG}
    \label{fig:experiment_open_closed_cilqg}
\end{figure}

We further compare our method with Open CILQG in this case, as shown in Fig.\ref{fig:experiment_open_closed_cilqg}. Samely, the first sub-figure shows the confidence ellipse of the planning results. We can see that CILQG can pass through the obstacles appropriately. On the other hand, although the Open CILQG result is collision-free, it is too conservative and cannot pass through the obstacles. This is because there is no closed-loop adjustment of the estimation of system uncertainty, and therefore it increases rapidly. From the speed profile, we can see the CILQG first decrease its velocity to pass through the obstacles safely and then accelerates to reach the reference speed. However, the Open CILQG  just conservatively stops before the obstacles. Nonetheless, both methods satisfy state chance-constraints according to the bottom sub-figure.

\begin{figure}
    \centering
    \includegraphics[width=0.45\textwidth]{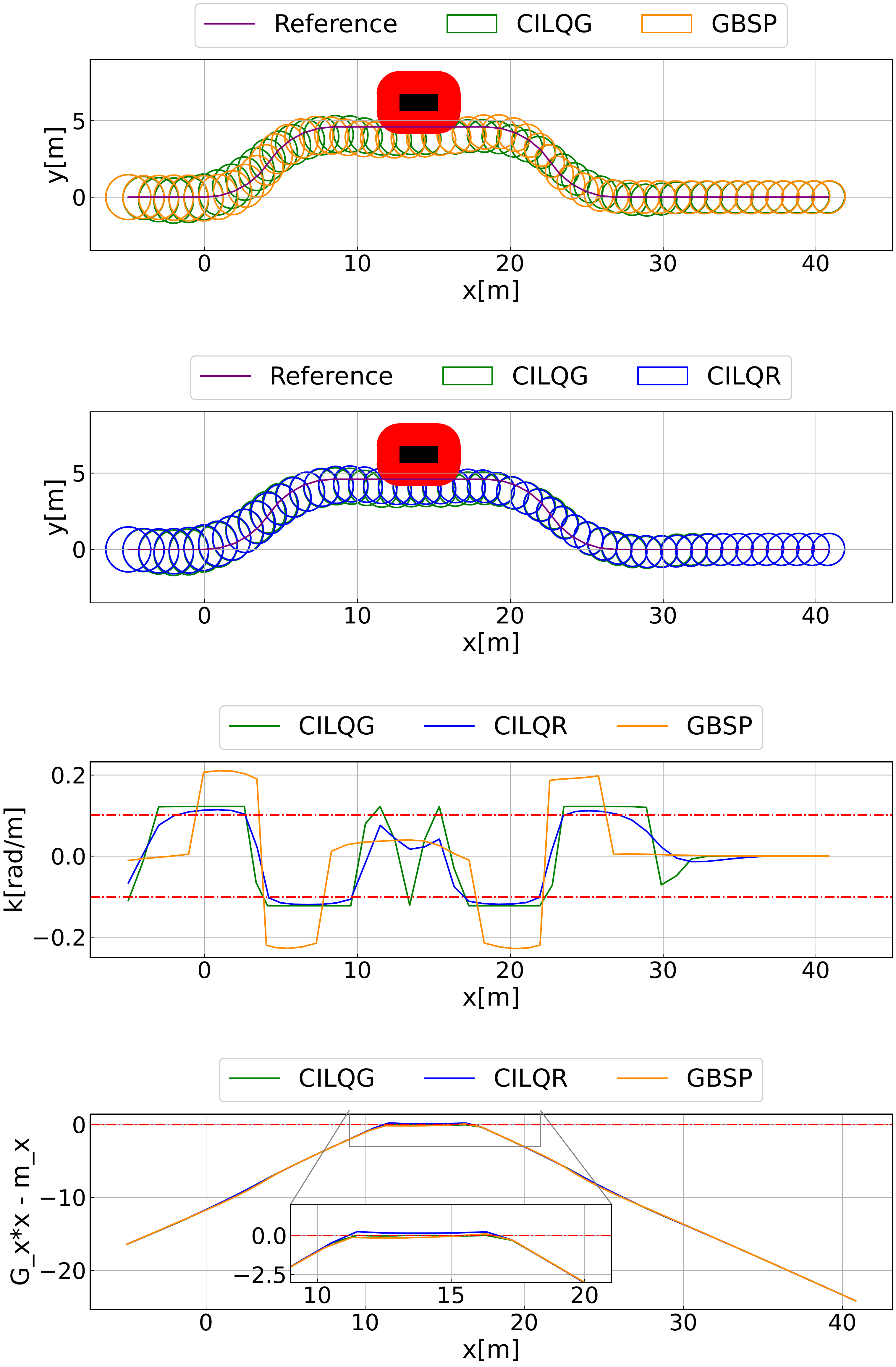}
    \caption{CILQG with one static obstacle}
    \label{fig:experiment_with_one_static_obstacle}
\end{figure}

Second, the autonomous vehicle is required to track a curved reference trajectory while avoiding a static obstacle. The result is shown in Fig.\ref{fig:experiment_with_one_static_obstacle}. The first two sub-figures show the confidence ellipse of the planning results, where only the CILQG result is collision-free. The third sub-figure shows the value of control input (curvature) applied on the vehicle. We can see that although all methods can track the curved reference, they apply significantly different values of control inputs. Only the CILQG result can satisfy the control input constraint, while others violate. The bottom sub-figure indicates that CILQG successfully handles state chance-constraint while other methods fail. Note that the red dashed line represents the constraint threshold on the curvature.

\subsection{Dynamic Obstacle Avoidance}
\begin{figure}
    \centering
    \includegraphics[width=0.45\textwidth]{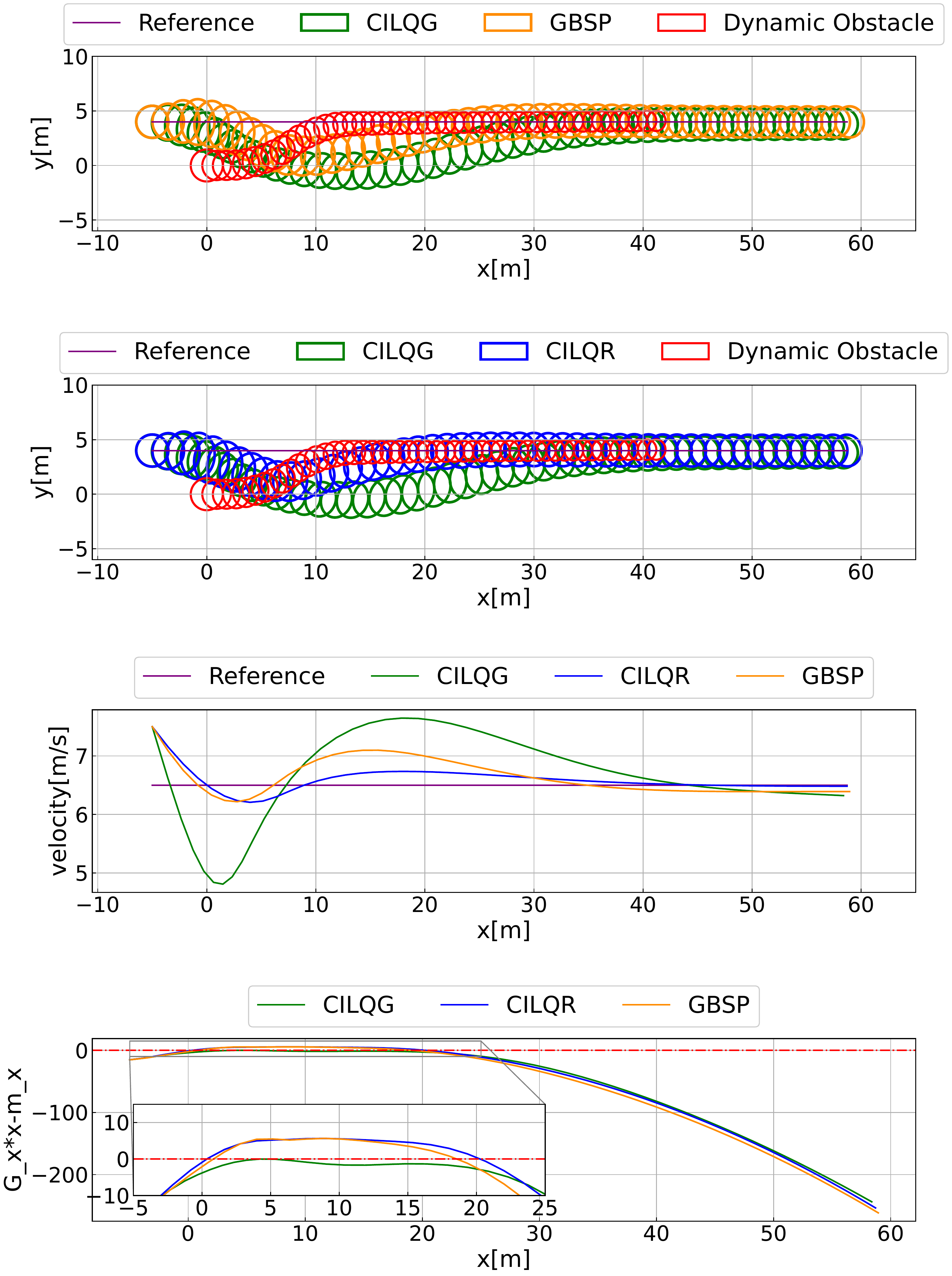}
    \caption{CILQG with dynamic obstacle}
    \label{fig:experiment_with_dynamic_obstacle}
\end{figure}

Next, we test our method in an environment with a dynamic obstacle running around. In this case, we use the constraints considering obstacle's uncertainty as described in Section~\ref{sec:uncertainty}. We assume a given stochastic trajectory of the obstacle with its mean and variance already specified. It can be obtained from an upstream motion prediction module of the autonomous driving system in practice. Fig.\ref{fig:experiment_with_dynamic_obstacle} shows the results, where the dynamic obstacle starts from the position $\left(0,0\right)$ and steers to track the reference trajectory, while the autonomous vehicle is required to overtake the obstacle from the downside. The first two sub-figures show the confidence ellipses of all methods, the third sub-figure shows the speed profile, and the last sub-figure shows the constraint violation. We can see that only our proposed CILQG method can guarantee safety while maintaining the efficiency of the trajectory.

\subsection{Runtime Analysis}
Finally, we evaluate the runtime of our proposed method for each experiment. We execute the algorithm 50 times and calculate the average time and its standard deviation. Table~\ref{tab:runtime} shows the summarized runtime. We can see that for the experiments with static obstacles, the computation time is strictly under 7ms. For the experiments with dynamic obstacles, the computation time is around 30ms. This is far enough for real-time autonomous driving as it usually requires a sampling time of 100ms~200ms for motion planning. Furthermore, the calculation efficiency can be further improved with better computation resources.
\begin{table}
  \begin{center}
    \caption{Runtime Analysis}
    \label{tab:runtime}
    \begin{tabular}{|c|c|c|} 
      \hline
      \textbf{Scenarios} & \textbf{mean [ms]} & \textbf{variance [ms]}\\
      \hline
      Two static obstacles & 5.56 & 0.93\\
      \hline
      One static obstacle & 4.26 & 0.69\\
      \hline
      Dynamic obstacle & 29 & 3.27\\
      \hline
    \end{tabular}
  \end{center}
\end{table}

\section{Conclusions}\label{sec:conclusions}
In this paper, we formulated a motion planning problem under uncertainties and stochastic constraints as a chance-constrained Gaussian belief space planning problem. We proposed the constrained iterative LQG (CILQG) algorithm to solve this problem. The proposed algorithm can find the optimal solution in real time by linearizing the system model iteratively to approximate the belief dynamics and transforming the original chance-constraints to standard linear constraints. Simulations for autonomous driving tasks in environments with both static and dynamic obstacles were conducted. Results indicated that CILQG is superior to baseline methods and had real-time computation efficiency.

\section*{ACKNOWLEDGMENT}
The author Y. Shimizu was supported by JST CREST GrantNumber JPMJCR19F3, Japan.

\bibliographystyle{ieee}
\bibliography{reference}

\end{document}